\documentclass[runningheads,a4paper]{llncs}

\usepackage{amssymb}
\usepackage{graphicx}
\usepackage{comment}
\usepackage{array}
\usepackage{textcomp}
\usepackage{enumerate}

\usepackage{xspace}

\def\ie{\textit{i.e.}~}

\def\etc{\emph{etc}.~} 
\def\vs{\emph{vs}.~}
\def\wrt{w.r.t. } 

\usepackage{color}

\usepackage{pifont} 
\newcommand{\cmark}{\ding{51}}
\newcommand{\xmark}{\ding{55}}

\usepackage{url}
\urldef{\mailsa}\path|{alfred.hofmann, ursula.barth, ingrid.haas, frank.holzwarth,|
	\urldef{\mailsb}\path|anna.kramer, leonie.kunz, christine.reiss, nicole.sator,|
	\urldef{\mailsc}\path|erika.siebert-cole, peter.strasser, lncs}@springer.com|    
\newcommand{\keywords}[1]{\par\addvspace\baselineskip
	\noindent\keywordname\enspace\ignorespaces#1}

\newcolumntype{L}[1]{>{\raggedright\let\newline\\\arraybackslash\hspace{0pt}}m{#1}}
\newcolumntype{C}[1]{>{\centering\let\newline\\\arraybackslash\hspace{0pt}}m{#1}}
\newcolumntype{R}[1]{>{\raggedleft\let\newline\\\arraybackslash\hspace{0pt}}m{#1}}

\usepackage{subfig}
\usepackage{booktabs}
\usepackage[hidelinks]{hyperref}

\begin{document}
	
	\mainmatter  
	
	\title{Capsule Networks against \\ Medical Imaging Data Challenges}
	
	\titlerunning{Capsule Networks against Medical Imaging Data Challenges}
	
	%
	\author{Amelia Jim\'{e}nez-S\'{a}nchez\inst{1}\href{https://orcid.org/0000-0001-7870-0603}{\includegraphics[scale=0.5]{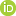}}, Shadi Albarqouni\inst{2}\href{https://orcid.org/0000-0003-2157-2211}{\includegraphics[scale=0.5]{images/orcid.png}}, Diana Mateus\inst{3}
	}
	\authorrunning{Capsule Networks against Medical Imaging Data Challenges}
	
	
	\institute{ 
		BCN MedTech, DTIC, Universitat Pompeu Fabra, Spain. \\
		\and
		Computer Aided Medical Procedures, Technische Universit\"{a}t M\"{u}nchen, Germany.
		\and 
		Laboratoire des Sciences du Num\'{e}rique de Nantes, UMR 6004, Centrale Nantes, France.
	}
	
	%
	%
	
	\toctitle{Lecture Notes in Computer Science}
	\tocauthor{Authors' Instructions}
	\maketitle
	
	\begin{abstract}
		
		A key component to the success of deep learning is the availability of massive amounts of training data. Building and annotating large datasets for solving medical image classification problems is today a bottleneck for many applications. Recently, capsule networks were proposed to deal with shortcomings of Convolutional Neural Networks (ConvNets).  
		In this work, we compare the behavior of capsule networks against ConvNets under typical datasets constraints of medical image analysis, namely, small amounts of annotated data and class-imbalance. We evaluate our experiments on MNIST, Fashion-MNIST and medical (histological and retina images) publicly available datasets. Our results suggest that capsule networks can be trained with less amount of data for the same or better performance and are more robust to an imbalanced class distribution, which makes our approach very promising for the medical imaging community.

		\keywords{capsule networks, small datasets, class imbalance.}
	\end{abstract}

	\section{Introduction}
	Currently, numerous state of the art solutions for medical image analysis tasks such as computer-aided detection or diagnosis rely on Convolutional Neural Networks (ConvNets)~\cite{geert2017survey}. The popularity of ConvNets relies on their capability to learn meaningful and hierarchical image representations directly from examples, resulting in a feature extraction approach that is flexible, general and capable of encoding complex patterns. However, their success depends on the availability of very-large databases representative of the full-variations of the input source. This is a problem when dealing with medical images as their collection and labeling are confronted with both data privacy issues and the need for time-consuming expert annotations. Furthermore, we have poor control of the class distributions in medical databases, \ie there is often an imbalance problem. Although strategies like transfer learning~\cite{zhou2017transflearn}, data augmentation~\cite{vasconcelos2017augmen} or crowdsourcing~\cite{albarqouni2016crowd} have been proposed, data collection and annotations is for many medical applications still a bottleneck~\cite{labels2017}.
	
	ConvNets' requirement for big amounts of data is commonly justified by a large number of network parameters to train under a non-convex optimization scheme. We argue, however, that part of these data requirements is there to cope with their poor modeling of spatial invariance. As it is known, purely convolutional networks are not natively spatially invariant. Instead, they rely on pooling layers to achieve translation invariance, and on data-augmentation to handle rotation invariance.  With pooling, the convolution filters learn the distinctive features of the object of interest irrespective of their location. Thereby losing the spatial relationship among features which might be essential to determine their class (e.g. the presence of plane parts in an image does not ensure that it contains a plane). 
	
	\begin{figure}[t]
		\centering
		{\includegraphics[width=1\textwidth]{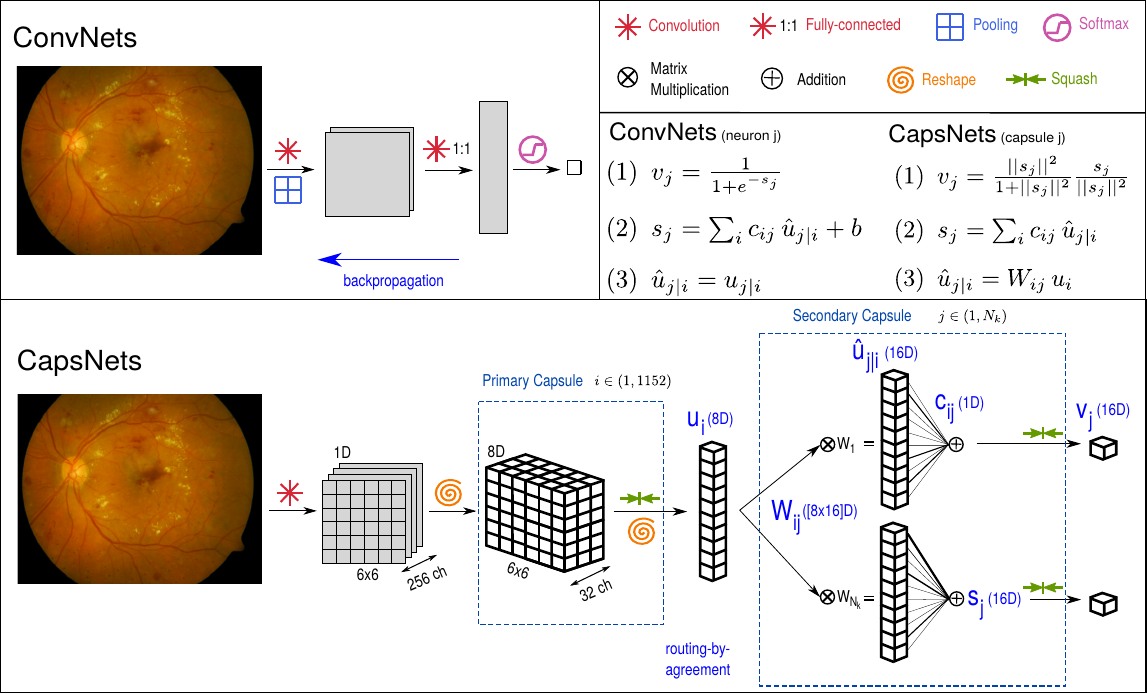}}
		\caption{Comparison of the flow and connections of ConvNets \vs CapsNets. Eq.~(1) shows the difference between the sigmoid and squashing functions. Eq.~(2) is a weighted sum of the inputs (ConvNets use bias). In CapsNets, $c_{ij}$ are the coupling coefficients. In (3), $\hat{u}_{j|i}$ is the transformed input to the \textit{j-th}~capsule/neuron. In CapsNets, the input from the \textit{i-th}~capsule is transformed with the weights $W_{ij}$. While in ConvNets, the raw input from the previous neuron is used.}
		\vspace{-0.55cm}
		\label{figure:diagram}
	\end{figure}
	
	Recently, capsule networks \cite{sabour2017capsnet} were introduced as an alternative deep learning architecture and training approach to model the spatial/viewpoint variability of an object in the image. Inspired by computer graphics, capsule networks not only learn good weights for feature extraction and image classification but also learn how to infer pose parameters from the image. Poses are modeled as multidimensional vectors whose entries parametrize spatial variations such as rotation, thickness, skewness, \etc As an example, a capsule network learns to determine whether a plane is in the image, but also if the plane is located to the left or right or if it is rotated. This is known as \textit{equivariance} and it is a property of human one-shot learning type of vision.
	
	In this paper, we experimentally demonstrate that the equivariance properties of CapsNets reduce the strong data requirements, and are therefore very promising for medical image analysis. Focusing on computer-aided diagnosis (classification) tasks, we address the problems of the limited amount of annotated data and imbalance of class distributions. To ensure the validity of our claims, we perform a large number of controlled experiments on two vision (MNIST and Fashion-MNIST) and two medical datasets that targets: mitosis detection (TUPAC16) and diabetic retinopathy detection (DIARETDB1).
	To the best of our knowledge, this is the first study to address data challenges in the medical image analysis community with Capsule Networks. 
	
	\section{Methods}
	In the following, we focus on the image classification problem characteristic of computer-aided diagnosis systems. Our objective is to study the behavior of Capsule Networks (CapsNets)~\cite{sabour2017capsnet} in comparison to standard Convolutional Networks (ConvNets) under typical constraints of biomedical image databases, such as a  limited amount of labeled data and class imbalance. We discuss the technical advantages that make CapsNets better suited to deal with the above-mentioned challenges and experimentally demonstrate their improved performance.
	
	\subsection{Capsule vs Convolutional Networks}
	
	Similar to ConvNet approaches, CapsNets build a hierarchical image representation by passing an image through multiple layers of the network. However, as opposed to the tendency towards deeper models, the original CapsNet is formed with only two layers:  a first \textit{primary caps} layer, capturing low-level cues, followed by a specialized \textit{secondary caps}, capable of predicting both the presence and \textit{pose} of an object in the image. The main technical differences of CapsNets \wrt ConvNets are:
	\\
	\indent\textit{i)} Convolutions are only performed as the first operation of the  \textit{primary caps} layer, leading as usual to a series of \textit{feature channels}.
	\\
	\indent \textit{ii)} Instead of applying a non-linearity to the scalar outputs of the convolution filters, CapsNets build tensors by grouping multiple feature channels (see the grid in Fig. \ref{figure:diagram}). The non-linearity, a \textit{squashing} function, becomes also a multidimensional operation, that takes the $j-th$ vector $s_j$ and restricts its range to the [0,1] interval to model probabilities while preserving the vector orientation. The result of the squashing function is a vector $v_j$, whose magnitude can be then interpreted as the probability of the presence of a capsule's entity, while the direction encodes its pose. $v_j$ is then the output of the capsule \textit{j}. 
	\\
	\indent\textit{iii)} The weights $W_{ij}$ connecting the $i$ primary capsule to the the $j-th$ secondary capsule are an affine transformation. These transformations allow learning part/whole relationships, instead of detecting independent features by filtering at different scales portions of the image.
	\\
	\indent\textit{iv)} The transformation weights $W_{ij}$ are not optimized with the regular backpropagation but with a \textit{routing-by-agreement} algorithm. The principal idea of the algorithm is that a lower level capsule will send its input to the higher level capsule that \textit{agrees} better with its input, this way is possible to establish the connection between lower- and higher-level information (refer to~\cite{sabour2017capsnet} for details). 
	\\
	\indent\textit{v)} Finally, the output  of a ConvNet is typically a softmax layer with cross-entropy loss:
	$
	\mathcal{L}_{ce} = - \sum_{x}g_l(x) \; log(p_l(x)) .
	$
	
	Instead, for every secondary capsule, CapsNet computes the margin loss for class $k$:
	\begin{equation}
	\mathcal{L}_k = T_k \; \max(0, m^{+} - ||\textbf{v}_k||)^2 + \lambda \; (1-T_k) \; \max(0, ||\textbf{v}_k|| - m^{-})^2,
	\end{equation}
	where the one-hot encoded labels $T_k$ are 1 iff an entity of class $k$ is present and $m^{+}=0.9$ and $m^{-}=0.1$, i.e. if an entity of class $k$ is present, its probability is expected to be above 0.9 ($||\textbf{v}_k||>0.9$), and if it is absent $||\textbf{v}_k||<0.1$. Since the threshold is not set as 0.5, the marginal loss forces the distances of the positive instances to be close to each other, resulting in a more robust classifier. The weight $\lambda=0.5$.
	
	As regularization method, CapsNet uses a decoder branch composed of two fully connected layers of 512 and 1024 filters respectively. The loss of this branch is the mean square error between the input image $x$ and its reconstruction $\hat{x}$ both of size $N\times M$,
	\begin{equation}
	\mathcal{L}_{\rm MSE} = \frac{1}{N \cdot M}\sum_{n=1}^{N}\sum_{m=1}^{M}{(x(n,m) - \hat{x}(n,m))^2} )
	\end{equation}
	The final loss, is a weighted average of the margin loss and the reconstruction loss
	$
	\mathcal{L}_{total} = \sum_{k=1}^{N_k} \mathcal{L}_k + \alpha \; \mathcal{L}_{MSE}.
	$
	
	\subsection{Medical Data Challenges}
	
	It is frequent for medical image datasets to be small and highly imbalanced. Particularly, for rare disorders or volumetric segmentation, healthy samples are the majority against the abnormal ones. The cost of miss-predictions in the minority class is higher than in the majority one since high-risk patients tend to be in the minority class. There are two common strategies to cope with such scenarios: i) increase the number of data samples and balance the class distribution, and ii) use weights to penalize stronger miss-predictions of the minority class. 
	
	We propose here to rely on the equivariance property of CapsNets to exploit the structural redundancy in the images and thereby reduce the number of images needed for training. 
	For example, in Fig.~\ref{figure:diagram}, we can see a fundus image in which diabetic retinopathy is present. There are different patterns present in the image that could lead to a positive diagnosis. Particularly, one can find soft and hard exudates or hemorrhages. While a ConvNet would tend to detect the presence of any of these features to make a decision, CapsNet routing algorithm is instead designed to learn to find relations between features. 
	Redundant features are collected by the routing algorithm instead of replicated in several parts of the network to cope with invariance. 
	We claim that the above advantages directly affect the number of data samples needed to train the networks.
	To demonstrate our hypothesis we have carefully designed a systematic and large set of experiments comparing a traditional ConvNet: LeNet \cite{lecun98lenet} and a standard ConvNet: Baseline from \cite{sabour2017capsnet}, against a Capsule Network \cite{sabour2017capsnet}. We focus on comparing their performance with regard to the medical data challenges to answer the following questions: 
	
	\begin{itemize}
		\item[$\circ$] How do networks behave under decreasing amounts of training data?
		\item[$\circ$] Is there a change in their response to class-imbalance?
		\item[$\circ$] Is there any benefit from data augmentation as a complementary strategy?
	\end{itemize}
	
	To study the generalization of our claims, our designed experiments are evaluated on four publicly available datasets for two vision and two medical applications:
	i) Handwritten Digit Recognition (MNIST), 
	ii) Clothes Classification (FASHION MNIST),
	iii) Mitosis detection,  a sub-task of mitosis counting, which is the standard way of assessing tumor proliferation in breast cancer images (TUPAC16 challenge \cite{tupac16web}), 
	and iv) Diabetic Retinopathy, an eye disease, that due to diabetes could end up in eye blindness over time. It is detected by a retinal screening test (DIARETDB1 dataset). Next, we provide some implementation details of the compared methods.

	\begin{table}[tb]
		\centering
		\scriptsize
		\begin{tabular}{|l|C{0.9cm}|C{0.9cm}|C{0.9cm}|C{0.9cm}|C{1.2cm}|C{1.2cm}|C{0.9cm}|C{0.9cm}|C{1cm}|C{1.3cm}|}
			\hline
			& Conv1 & Pool1 & Conv2 & Pool2 & Conv3 & - & FC1 & Drop & FC2 & \#Params.\\
			\hline  
			LeNet & \begin{tabular}{@{}c@{}} $5\times5$ \\ 6 ch \end{tabular} & \begin{tabular}{@{}c@{}} $2\times2$ \\ \phantom{0} \end{tabular} & \begin{tabular}{@{}c@{}} $5\times5$ \\ 16 ch \end{tabular} & 
			\begin{tabular}{@{}c@{}} $2\times2$ \\ \phantom{0} \end{tabular}&
			\xmark & - & \begin{tabular}{@{}c@{}} $1\times1$ \\ 120 ch \end{tabular} & \xmark & 
			\begin{tabular}{@{}c@{}} $1\times1$ \\ 84 ch \end{tabular} & $ 60K$ \\
			\hline
			Baseline & \begin{tabular}{@{}c@{}} $5\times5$ \\ 256 ch \end{tabular} & \xmark & \begin{tabular}{@{}c@{}} $5\times5$ \\ 256 ch \end{tabular} & 
			\xmark & \begin{tabular}{@{}c@{}} $5\times5$ \\ 128 ch \end{tabular} & - & \begin{tabular}{@{}c@{}} $1\times1$ \\ 328 ch \end{tabular} & \cmark & 
			\begin{tabular}{@{}c@{}} $1\times1$ \\ 192 ch \end{tabular} & $ 35.4M$ \\
			\hline
			\hline
			& Conv1 & Pool1 & Conv2 & Pool2 & Caps1 & Caps2 & FC1 & Drop & FC2 & \#Params. \\
			\hline 
			CapsNet & \begin{tabular}{@{}c@{}} $9\times9$ \\ 256 ch \end{tabular} & \xmark & \begin{tabular}{@{}c@{}} $9\times9$ \\ 256 ch \end{tabular} & \xmark & \begin{tabular}{@{}c@{}} 1152 caps \\ 8D \end{tabular} & \begin{tabular}{@{}c@{}} $N_k$ caps \\ 16D \end{tabular} & \begin{tabular}{@{}c@{}} $1\times1$ \\ 512 ch \end{tabular} & \xmark & \begin{tabular}{@{}c@{}} $1\times1$ \\ 1024 ch \end{tabular}&8.2M\\
			\hline
		\end{tabular}
		\caption{Details of each of the architectures. For convolution, we specify the size of the kernel and the number of output channels. In the case of pooling, the size of the kernel. And for capsule layers, first, the number of capsules and, in the second row, the number of dimensions of each capsule. } 
		\vspace{-0.8cm}
		\label{table:architectures}
	\end{table}
	
	\paragraph{Architectures} 
	Since research of capsules is still in its infancy, we pick the first ConvNet, LeNet~\cite{lecun98lenet} for a comparison. Though this network has not many parameters (approx. 60K), it is important to notice the presence of pooling layers which reduce the number of parameters and lose the spatial relationship among features. For a fairer comparison, we pick another ConvNet with similar complexity to CapsNet, in terms of training time, that has no pooling layers, which we name hereafter Baseline and was also used for comparison in \cite{sabour2017capsnet}.
	
	\textbf{LeNet} has two convolutional layers of 6 and 16 filters. Kernels are of size 5x5 and stride 1. Both are followed by a ReLU and pooling of size 2x2. Next, there are two fully connected layers with 120 and 84 filters. \textbf{Baseline} is composed of three convolutional layers of 256, 256, 128 channels, with 5x5 kernel and stride of 1. Followed by two fully connected layers of size 382, 192 and dropout. In both cases, the last layer is connected to a softmax layer with cross-entropy loss.
	For \textbf{CapsNet \cite{sabour2017capsnet}}, we consider two convolutional layers of 256 filters with kernel size of 9x9 and stride of 1. Followed by two capsule layers of 8 and 16 dimensions, respectively, as depicted in Fig.~\ref{figure:diagram}. For each of the 16-dimensional vectors that we have per class, we compute the margin loss like \cite{sabour2017capsnet} and attach a decoder to reconstruct the input image. Details are summarized in Table~\ref{table:architectures}.
	
	\vspace{-.3cm}
	\paragraph{Implementation.}
	The networks were trained on a Linux-based system, with 32 GB RAM, Intel(R) Core(TM) CPU @ 3.70 GHz and 32 GB GeForce GTX 1080 graphics card. All models were implemented using Google’s Machine Learning library TensorFlow \footnote{https://www.tensorflow.org/}. The convolutional layers are initialized with Xavier weights \cite{glorot2010weights}. All the models were trained
	in an end to end fashion, with Adam optimization algorithm \cite{kingm12014adam}, using grayscale images of size $28 \times 28$. The batch size was set to 128. For MNIST and Fashion-MNIST, we use the same learning rate and weight for the reconstruction loss as \cite{sabour2017capsnet}, while for AMIDA and DIARETDB1 we reduced both by 10. If not otherwise stated, the models were trained for 50 epochs. The reported results were tested at minimum validation loss.
	
	\section{Experimental Validation}
	
	Our systematic experimental validation compares the performance of LeNet, a Baseline ConvNet and CapsNet with regard to the three mentioned data-challenges, namely the limited amount of training data, the class-imbalance, and the utility of data-augmentation. We trained in total 432 networks, using 3 different architectures, under 9 different data conditions, for 4 repetitions, and for 4 publicly available datasets.
	The two first datasets are the well known MNIST \cite{lecun2010mnist} and Fashion-MNIST \cite{xiao2017fashion}, with 10 classes and, 60K and 10K images for training and test respectively.  
	
	For \textit{mitosis detection}, we use the histological images of the first auxiliary dataset from the TUPAC16 challenge \cite{tupac16web}. There are a total of 73 breast cancer images, of $2K \times 2K$ pixels each, and with the annotated location coordinates of the mitotic figures. Images are normalized using color deconvolution~\cite{vahadane2016stain} and only the hematoxylin channel is kept. We extract patches of size $100 \times 100$ pixels that are downsampled to $28 \times 28$, leading to about 60K and 8K images for training and test respectively. The two classes are approximately class-wise balanced after sampling. 
	
	For the \textit{diabetic retinopathy detection}, we consider DIARETDB1 dataset \cite{kalesnykieneretina}. It consists of 89 color fundus images of size $1.1K \times 1.5K$ pixels, of which 84 contain at least mild signs of the diabetic retinopathy, and 5 are considered as normal. Ground truth is provided as masks. We enhance the contrast of the fundus images by applying contrast limited adaptive histogram equalization (CLAHE) on the lab color space and keep only the green channel. We extract patches of $200 \times 200$ pixels that are resized to $28 \times 28$. This results in about 50K and 3K images for training and test respectively. They are approximately class-wise balanced after sampling.
	
	\subsection{Limited amount of training data}
	
	We compare the performance of the two networks for the different classification tasks when the original amount of training data is reduced to $50\%, 10\%, 5\%$, and $1\%$ while keeping the original class distribution.
	We run each of the models for the same number of iterations that are required to train 50 full epochs using all the training data. Early-stop is applied if the validation loss does not improve in the last 20 epochs.
	
\begin{figure}[t]
		\scriptsize
		\center
		\includegraphics[width=7.5cm]{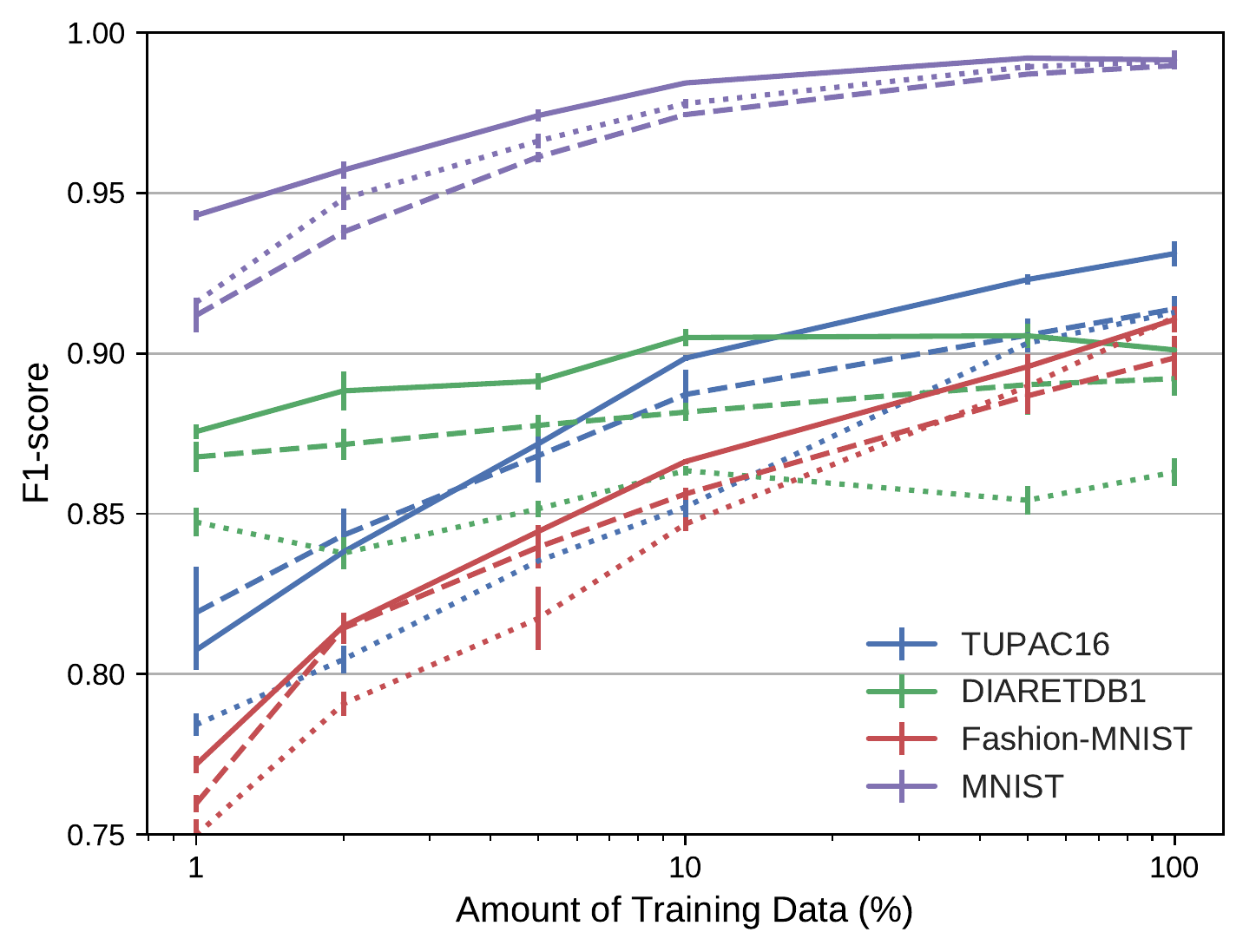} 
		\caption{Mean $F_1$-score and standard deviation (4 runs) for different amounts of training data. Solid line: CapsNet, dotted line: Baseline, and dashed line: LeNet.} 
		\vspace{-0.5cm}
		\label{fig:train-size} 
	\end{figure}

	The  results are shown in Table~\ref{table:exp-train-size}. For almost all scenarios CapsNet performs better than LeNet and Baseline. We can observe in Figure \ref{fig:train-size} how for MNIST the gap is higher for a small amount of data and is reduced when more data is included. LeNet with 5\% of the data has a similar performance to CapsNet, and better than Baseline, with 1\% of the data for DIARETDB1. We attribute this behavior to the structures that are present in this type of images. All the experiments validated the significance test with a p-value \textless{}
	0.05, except for those on the TUPAC16 dataset, we presume this is associated to the CapsNet limitations that we present in Section \ref{sec:conclusions}.
	
	\begin{table}[tb]
		\centering
		\scriptsize
		\subfloat[][Mean $F_1$-score using \textbf{different amounts of training data}.]
		{
			\begin{tabular}{l|C{0.8cm}|C{0.8cm}|C{1.05cm}|C{0.8cm}|C{0.8cm}|C{1.05cm}|C{0.8cm}|C{0.8cm}|C{1.05cm}|C{0.8cm}|C{0.8cm}|C{1.05cm}|C{0.8cm}|C{0.8cm}|C{1.05cm}|}  
				Training Data & \multicolumn{3}{c}{1\%} & \multicolumn{3}{|c}{5\%} & \multicolumn{3}{|c}{10\%} & \multicolumn{3}{|c|}{50\%} \\
				\hline
				& LeNet & Base. & CapsNet & LeNet & Base. & CapsNet & LeNet & Base. & CapsNet & LeNet & Base. & CapsNet \\
				\hline
				TUPAC16 & \bf{0.822} & 0.784 & 0.809 & 0.872 & 0.835 & 0.872 & 0.890 & 0.852 & \bf{0.898} & 0.908 & 0.903 & \bf{0.923} \\
				DIARETDB1 & 0.870 & 0.847 & \bf{0.875} & 0.877 & 0.852 & \bf{0.893} & 0.883 & 0.863 & \bf{0.907} & 0.895 & 0.854 & \bf{0.908}  \\
				Fashion-M. & 0.759 & 0.749 & \bf{0.772} & 0.841 & 0.817 & \bf{0.846} & 0.856 &  0.847 & \bf{0.866} & 0.885 & 0.889 & \bf{0.896}  \\ 
				MNIST & 0.909 & 0.916 & \bf{0.943} & 0.961 & 0.966 & \bf{0.975} & 0.975 & 0.978 & \bf{0.985} & 0.987 & 0.989 & \bf{0.992}  \\
			\end{tabular}
			\label{table:exp-train-size}
		} \\
		\subfloat[][Mean $F_1$-score reported for different \textbf{class-imbalance} scenarios.]
		{
			\begin{tabular}{l|C{1.2cm}|C{1.2cm}|C{1.2cm}|C{1.2cm}|C{1.2cm}|C{1.2cm}|C{1.2cm}|C{1.2cm}|C{1.2cm}|}
				Scenario & \multicolumn{3}{c}{Balanced} & \multicolumn{3}{|c}{Imbalanced 1} & \multicolumn{3}{|c|}{Imbalanced 2}  \\
				\hline
				& LeNet & Baseline & CapsNet & LeNet & Baseline & CapsNet & LeNet & Baseline & CapsNet  \\
				\hline
				TUPAC16 & 0.914 & 0.913 & \bf{0.932} & 0.881 & 0.813 & \bf{0.892} & 0.905 & 0.874 & \bf{0.909}\\
				DIARETDB1 & 0.895 & 0.863 & \bf{0.899} & 0.869 & 0.839 & \bf{0.887} & 0.889 & 0.874 & \bf{0.898}\\
				Fashion-M. & 0.899 & \bf{0.911} & 0.910 & 0.890 & \bf{0.902} & 0.889 & 0.871 & \bf{0.881} & 0.863 \\ 
				MNIST & 0.989 & 0.991 & 0.991 & 0.988 & 0.989 & \bf{0.993} & 0.985 & 0.987 & \bf{0.992} \\
			\end{tabular}
			\label{table:exp-imbalanced}
		}
		\\ 
		\subfloat[][Mean $F_1$-score with and wihtout \textbf{data augmentation}.]
		{
			\begin{tabular}{l|C{1.2cm}|C{1.2cm}|C{1.2cm}|C{1.2cm}|C{1.2cm}|C{1.2cm}|}
				Data Augmentation & \multicolumn{3}{c}{No} & \multicolumn{3}{|c|}{Yes}   \\
				\hline
				& LeNet & Baseline & CapsNet & LeNet & Baseline & CapsNet \\
				\hline
				TUPAC16 & 0.904 & 0.892 & \bf{0.914} & 0.914 & 0.913 & \bf{0.932}  \\
				DIARETDB1 & 0.883 & 0.864 & \bf{0.895} & 0.892 & 0.863 & \bf{0.899} \\
				Fashion-MNIST & 0.899 & \bf{0.911} & 0.910 & 0.902 & 0.911 & \bf{0.913} \\ 
				MNIST & 0.989 & 0.991 & 0.991 & 0.990 & 0.993 & \bf{0.994}   \\
			\end{tabular}
			\label{table:exp-augmentation}
		}
		\caption{F-1 scores under different data-challenges.} \label{table:full}
		\vspace{-0.6cm}
	\end{table}

	\subsection{Class-imbalance}
	For the medical datasets, we simulate class imbalance by reducing to $20\%$ one of the two classes. Initially, we reduce abnormal class and, afterward, the healthy class. For the other two datasets, we decrease two classes at the same time. For MNIST, we first consider reducing the classes ``0" and ``1" and secondly, the classes ``2" and ``8". Similar for Fashion-MNIST, we reduce the classes ``T-shirt/top" and ``Trouser", and in the second scenario, ``Pullover" and ``Shirt".
	
	In Table \ref{table:exp-imbalanced} results are reported. Again, CapsNet surpasses the performance of ConvNets for all cases, except for Fashion-MNIST where the f1-scores are similar. At least one of the imbalance cases verified the significance test for all datasets.
	
	\subsection{Data augmentation}
	
	In the last series of experiments, we compare the performance of the three networks using data augmentation, a common technique to increase the amount of training data and balance class distributions. The original dataset is augmented with  $\pm 10$ degrees rotations, with a translation of $\pm 30$ pixels for medical datasets, and with flips (horizontal for Fashion-MNIST and, both horizontal and vertical for TUPAC16 and DIARETDB1). MNIST and Fashion-MNIST are augmented by 5\%, for the other two datasets we consider the no augmented version to be 50\% (TUPAC16) and 90\% (DIARETDB1) smaller.
	
	The performances in Table \ref{table:exp-augmentation} show that, CapsNet \textit{without} data augmentation achieves a similar (TUPAC16, MNIST, Fashion-MNIST) or even better (DIARETDB1) performance than ConvNets using data augmentation. All results are significant, the only Baseline for MNIST is comparable to the performance of CapsNet. These results confirm the benefits of equivariance over invariance.
	
	\section{Conclusion}
	\label{sec:conclusions}
	In this work, we experimentally demonstrate the effectiveness of using CapsNet to improve CADx classification performance under medical data challenges. In particular, we demonstrate the increased generalization ability of CapsNets \vs ConvNets when dealing with the limited amount of data and class-imbalance. The performance improvement is a result of CapsNets equivariance modeling, that is, its ability to learn pose parameters along with filter weights. Together with the \textit{routing-by-agreement} algorithm, this paradigm change requires to see fewer viewpoints of the object of interest, and therefore fewer images, in order to learn the discriminative features to classify them.
	We have also reported limitations to this otherwise general improvement of CapsNets over ConvNets, their improvement in performance is significant but has a limit that we observed for the more complex TUPAC dataset at 1\% (5.5K training samples).
	
	Classification tasks where the global spatial structure plays a role can better exploit the advantages of CapsNets (DIARETDB1). 
	
	One of the disadvantages of routing-by-agreement is that is slower than regular backpropagation, CapsNet with 8.2M parameters take about the same training time per epoch than Baseline with 35.4M (a ResNet-50 has 25.6M parameters). These architectures lack purposed layers, e.g. batch normalization, that could help to ease the convergence. Depending on the number of classes, CapsNet and Baseline need between 1-3 minutes per epoch, while LeNet runs in 1-2 seconds.
	
\begin{figure}[t]
		\scriptsize
		\center
		\includegraphics[width=10cm]{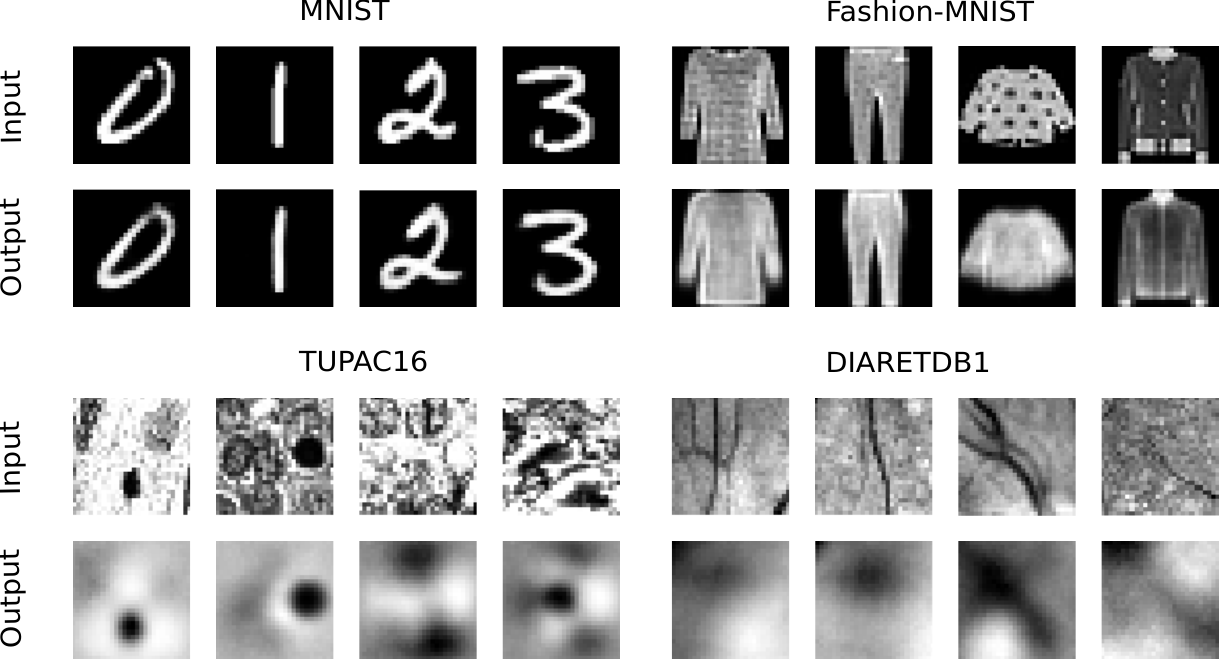} 
		\caption{Test input images and their reconstructions.} 
		\vspace{-0.5cm}
		\label{fig:reconstructions} 
	\end{figure}

	Also, when visualizing the images reconstructed through the encoder-decoder branch (Fig.~\ref{fig:reconstructions}), we observe that they are blurry, especially for medical datasets with complex backgrounds. The fully-connected layers of this branch seem to be good enough to regularize the parameter optimization but lose a lot of information. Our future work includes replacing these layers with deconvolutions to get a better insight into the learned latent space.
	
	We recommend the use of capsule networks for medical datasets where the structure is important and patterns appear in different parts of the input images, as it is for retina. Our results confirm that they perform better than standard ConvNets for the limited amount of data, at least of the order of 10k. Another potential application would be the detection of rare diseases or segmentation due to the high performance under class-imbalance.
	
	\subsubsection*{Acknowledgment.} 
	This work has received funding from the European Union’s Horizon 2020 research and innovation programme under the Marie Sklodowska-Curie grant agreement No. 713673. Amelia Jim\'{e}nez-S\'{a}nchez has received financial support through the ``la Caixa'' INPhINIT Fellowship Grant for Doctoral studies at Spanish Research Centres of Excellence, ``la Caixa'' Banking Foundation, Barcelona, Spain. The authors would like to thank Nvidia for the GPU donation and Aur\'{e}lien Geron for his tutorial and code on Capsule Networks.

	\bibliographystyle{splncs03}
	\bibliography{biblio}

\end{document}